\documentclass{article}
\pdfoutput=1
\usepackage[final]{corl_2023} 

\usepackage{times}

\usepackage[numbers]{natbib}
\usepackage{multicol}

\usepackage{graphicx}

\usepackage{amsmath}
\usepackage{amssymb}
\usepackage{booktabs}
\let\labelindent\relax
\usepackage{enumitem}
\usepackage{tabularx}
\usepackage{xcolor}
\usepackage{xspace}

\usepackage{caption} 
\usepackage{subcaption}

\makeatletter
\def\blfootnote{\gdef\@thefnmark{}\@footnotetext}
\makeatother

\usepackage{etoc}
\usepackage{subfiles}

\newcommand{\model}{RoboCook\xspace}

\title{\model: Long-Horizon Elasto-Plastic Object Manipulation with Diverse Tools}

\author{
  Haochen Shi$^{\textbf{1*}}$
  \And
  Huazhe Xu$^{\textbf{1*{\dag}}}$
  \And
  Samuel Clarke$^{\textbf{1}}$
  \And
  Yunzhu Li$^{\textbf{1,2}}$
  \And
  Jiajun Wu$^{\textbf{1}}$
  \and
  $^{1}$Stanford University
  \and
  $^{2}$UIUC
  \and
  *Equal contribution
  \and
  \url{https://hshi74.github.io/robocook}
}

\begin{document}
\maketitle

\vspace{-5mm}
\begin{abstract}
Humans excel in complex long-horizon soft body manipulation tasks via flexible tool use: bread baking requires a knife to slice the dough and a rolling pin to flatten it. Often regarded as a hallmark of human cognition, tool use in autonomous robots remains limited due to challenges in understanding tool-object interactions. 
Here we develop an intelligent robotic system, \model, which perceives, models, and manipulates elasto-plastic objects with various tools. \model uses point cloud scene representations, models tool-object interactions with Graph Neural Networks (GNNs), and combines tool classification with self-supervised policy learning to devise manipulation plans. We demonstrate that from just 20 minutes of real-world interaction data per tool, a general-purpose robot arm can learn complex long-horizon soft object manipulation tasks, such as making dumplings and alphabet letter cookies. Extensive evaluations show that \model substantially outperforms state-of-the-art approaches, exhibits robustness against severe external disturbances, and demonstrates adaptability to different materials. 



\end{abstract}



\blfootnote{{\dag}~Now at Institute for Interdisciplinary Information Sciences, Tsinghua University, Beijing, China}
\section{Introduction}

Think about all the steps and tools a robot would need to use to make a dumpling from a lump of dough. This scenario contains three fundamental research problems in robotics: deformable object manipulation~\citep{matas2018sim, lin2021softgym, zhu2022challenges, yin2021modeling}, long-horizon planning~\citep{hartmann2022long, nairhierarchical, pirk2021modeling, simeonov2021long}, and tool usage~\citep{billard2019trends, yamanobe2017brief, seitatoolflownet, Finn-RSS-19}. The task poses significant challenges to the robot, because it involves decisions at both discrete (e.g., which tool to use) and continuous levels (e.g., motion planning conditioned on the selected tool).


\begin{figure}[t!]
\centering
\includegraphics[width=0.98\columnwidth]{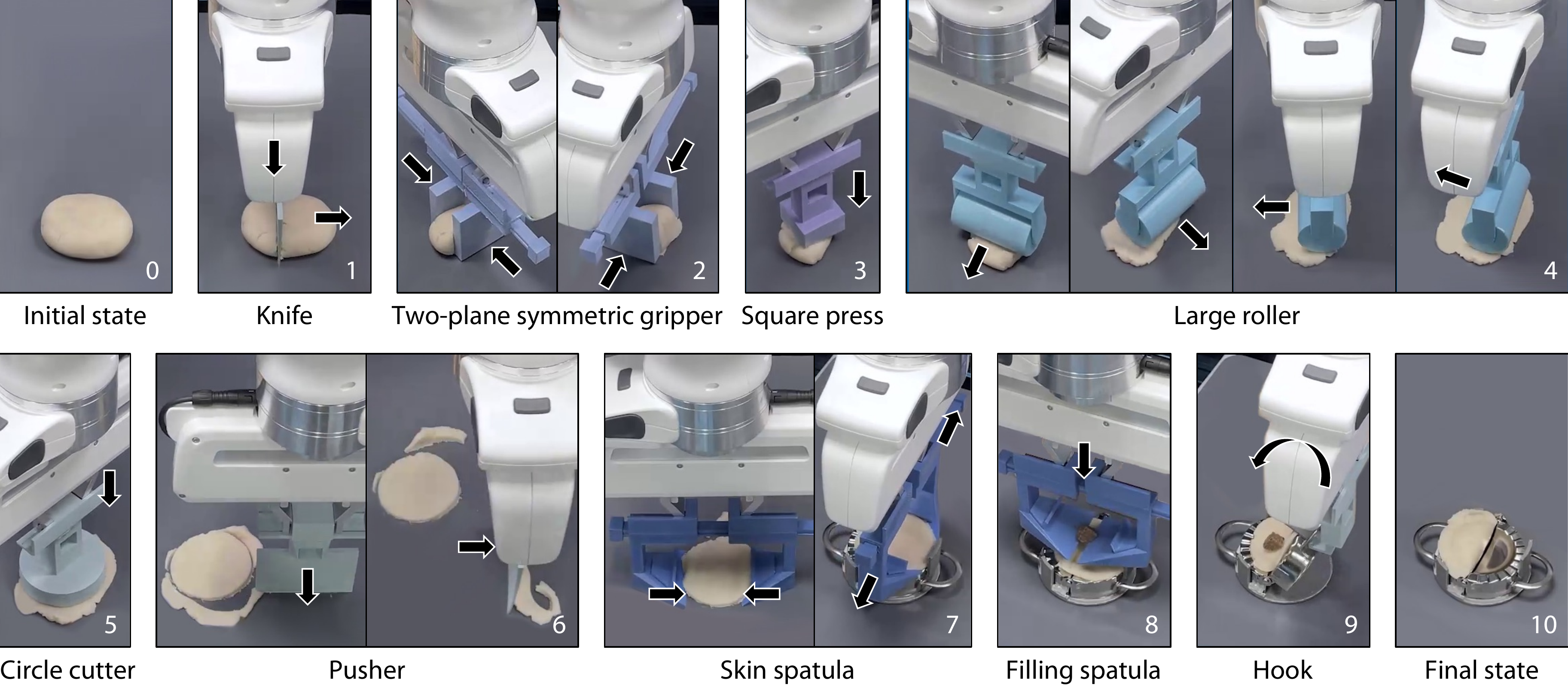}
\caption{\textbf{Making dumplings.}~\model makes a dumpling from a piece of dough in nine steps: The robot (1) cuts the dough to an appropriate volume, (2) pinches the dough and regularizes the shape, (3) presses to flatten the dough, (4) rolls to flatten the dough further, (5) cuts a circular dumpling skin, (6) removes the excess dough, (7) picks and places the skin onto the mold, (8) adds the filling, and (9) closes and opens the mold. The black arrows denote the moving direction.}
\label{fig:dumpling}
\vspace{-2mm}
\end{figure}

To address these challenges, we propose \model, a framework that perceives, models, and manipulates elasto-plastic objects for long-horizon tasks like making dumplings and alphabet letter cookies. \model introduces three technical innovations. First, we apply a data-driven approach with a Graph Neural Network (GNN)~\citep{scarselli2008graph, li2018learning, li2019propagation} to learn highly complex interactions between the soft object and various tools purely from visual observations. Second, we combine a PointNet-based \citep{qi2017pointnet, qi2017pointnet++} tool classification module with learned dynamics models to determine the most appropriate tool to use at the current task stage. Third, we use a self-supervised policy trained with synthetic data generated by our learned dynamics model for gripping, rolling, and pressing to improve performance and speed, and hand-coded policies for other skills. 

We carry out comprehensive evaluations to show \model's effectiveness, robustness, and generalizability. Figure~\ref{fig:dumpling} shows a typical successful trial of making a dumpling. To showcase robustness, we apply external perturbations during real-time execution, and \model still succeeds in making a dumpling. To demonstrate generalizability, we test \model to make alphabet letter cookies and \model outperforms four strong baselines by a significant margin. \model can also generalize to various materials by making a particular shape on different materials without retraining.


\section{Related Work}

\textbf{Long-horizon deformable object manipulation.}~Real-world deformable object manipulation is a challenging task~\citep{Shi-RSS-22, matl2021deformable, lin2022planning}. The innate high DoFs, partial observability, and non-linear local interactions make deformable objects hard to represent, model, and manipulate~\citep{driess2023learning, Chi-RSS-22, ha2022flingbot}. Conventional approaches in deformable object manipulation choose model-based methods~\citep{huang2020plasticinelab, lin2021diffskill, cretu2011soft, li2021contact} or adaptive methods~\citep{navarro2016automatic, cherubini2020model, yoshimoto2011active}, yielding complex system identification problems. For example, \citet{li2021contact} proposes a differentiable simulator as the surrogate of real-world elasto-plastic objects. However, these simulators are based on approximate modeling techniques, resulting in a noticeable sim-to-real gap that hinders their application to real-world scenarios
. There have been efforts toward learning to manipulate from expert demonstrations~\citep{balaguer2011combining, nadon2018multi, jia2019cloth}. This paradigm is used in manipulating liquid~\citep{seitatoolflownet}, sand~\citep{cherubini2020model}, and dough~\citep{figueroa2016learning}. Despite these successes, obtaining the demonstration is expensive and sometimes prohibitive. Another recent trend is to learn a dynamics model from high-dimensional sensory data directly for downstream manipulation tasks~\citep{Shi-RSS-22, battaglia2016interaction, chang2016compositional, sanchez2018graph, li2019propagation, silver2021planning}. While these approaches show impressive results, they only consider short-horizon tasks using just one tool. In contrast, long-horizon tasks like dumpling-making require a reactive planner to understand the long-term physical effect of different tools to make the most effective discrete (e.g., which tool to use) and continuous (e.g., the action parameters) action decisions.

\textbf{Tool usage.}~Tool usage is widely studied in cognitive science and robotics research~\citep{van2003model, holladay2019force}. To study the process of human evolution, many researchers endorse tool usage as a main benchmark to evaluate the intelligence of living primates with respect to that of extinct hominids~\citep{parker1977object, ingold1993tool, matsuzawa2008primate}. To equip robots with the same capability of tool usage as humans, prior works focus on teaching the robot the representation and semantics of tools or objects that potentially function like tools for downstream policy learning~\citep{seitatoolflownet, Finn-RSS-19}. To perform complicated long-horizon tasks composed of several subtasks, prior works also use human demonstrations to learn a hierarchical policy network~\citep{liang2022learning}. In this work, we will use real-world robot random-play data, which is cheaper to acquire than human demonstration data. We train the robot on these self-exploratory trials to understand the physical interactions between different tools and deformable objects.

\section{Method}

The \model framework has three major components: perception, dynamics, and closed-loop control. We first use an effective sampling scheme and intuitive tool representations for particle-based scene representation. Second, we train Graph Neural Networks (GNNs) as the dynamics model from the processed video dataset to accurately predict dough states during manipulation. Last, a tool classifier selects the best tool for each substage in a long-horizon task and a self-supervised policy network performs closed-loop control.

\subsection{Perception}
\begin{figure}[t!]
\centering
\includegraphics[width=\columnwidth]{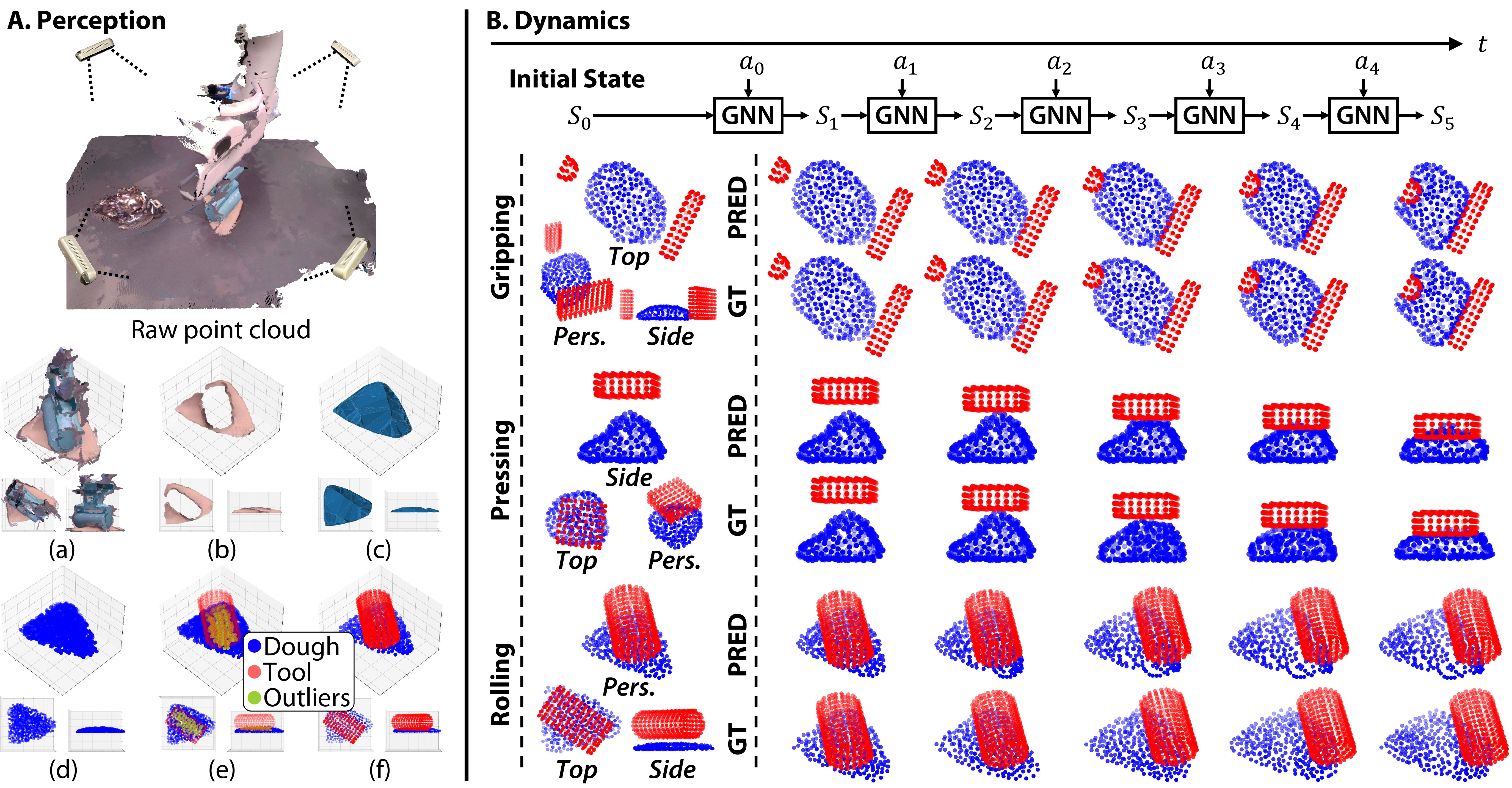}
\caption{\textbf{Perception and dynamics of \model}.~\textbf{(A)} The input to the perception module is a point cloud of the robot’s workspace captured by four RGB-D cameras. From the raw point cloud, we (a) crop the region of interest, (b) extract the dough point cloud, (c) reconstruct a watertight mesh (d) use the Signed Distance Function (SDF) to sample points inside the mesh, (e) remove points within the tools' SDF, and (f) sample 300 surface points. \textbf{(B)} We process videos of each tool manipulating the dough into a particle trajectory dataset to train our GNN-based dynamics model. The model can accurately predict long-horizon state changes of the dough in gripping, pressing, and rolling tasks. On the left are the initial states' perspective, top, and side views, and on the right is a comparison of model predictions and ground truth states.}
\label{fig:dynamics}
\vspace{-0.1em}
\end{figure}

The perception module aims to sample particles sparsely and uniformly for the downstream dynamics model. This task is challenging because of dough occlusions from the robot and tool and self-occlusions from the irregular and concave shape of the dough.

We merge point clouds from four calibrated RGB-D cameras and perform color segmentation to extract the dough point cloud. Then we apply either Poisson surface reconstruction~\citep{kazhdan2006poisson} or alpha-shape surface reconstruction~\citep{edelsbrunner1994three} to reconstruct a watertight surface of the dough, depending on occlusion levels. In heavy occlusion cases, alpha-shape surface reconstruction is usually worse at capturing concavities than Poisson surface reconstruction. However, we use it to secure a complete and singular mesh. With few or no occlusions, we combine Poisson surface reconstruction and the MeshFix algorithm~\citep{attene2010lightweight} to generate a watertight mesh. We use the watertight mesh's Signed Distance Function (SDF) to sample points inside it randomly. To compensate for details lost during surface reconstruction, we apply a voxel-grid filter to reconstruct concavities by removing the points above the original dough point cloud. We also remove the noisy points penetrating the tool using the mesh's SDF. We compute the tool mesh's SDF from the robot’s end-effector pose (recorded during data collection) and the ground-truth tool mesh. We then perform alpha-shape surface reconstruction and use Poisson disk sampling~\citep{yuksel2015sample} to sample 300 points uniformly on the surface, capturing more details of the dough with a fixed particle number.


For different tools, we uniformly sample particles on the surface of their ground truth mesh to reflect their geometric features. The complete scene representation for the downstream dynamics model concatenates dough and tool particles. Section~\ref{data_collection} and~\ref{data_preprocessing} of supplementary materials describe more data collection and preprocessing details.

\subsection{Dynamics Model}

Our GNN-based dynamics model predicts future states of dough based on the current state and actions of the tools with only 20 minutes of real-world data per tool. The rigid and non-rigid motions are each predicted by a multi-layer perceptron (MLP) and added together as the predicted motion by the GNN. 

The graph of sampled particles at each time step is represented as ${s_t=\left({\mathcal O}_t, {\mathcal E}_t\right)}$ with ${\mathcal O}_t$ as vertices, and ${\mathcal E}_t$ as edges. For each particle, ${{\mathbf o}_{i,t}=\left({\mathbf x}_{i,t}, \mathbf c_{i,t}^{o}\right)}$, where ${\mathbf x}_{i,t}$ is the particle position $i$ at time $t$, and $\mathbf c_{i,t}^{o}$ is the particle's attributes at time $t$, including the particle normal and group information (i.e., belongs to the dough or the tool). The edge between a pair of particles is denoted as ${{\boldsymbol e}_k=\left(u_k,v_k\right)}$, where $1 \leq u_k$, $v_k \leq \left|{\mathcal O}_t\right|$ are the receiver particle index and sender particle index respectively, and $k$ is the edge index. Section~\ref{graph_building} of supplementary materials provides more details on graph building.


Following the previous work~\citep{Shi-RSS-22}, we use a weighted function of Chamfer Distance (CD)~\citep{fan2017point} and Earth Mover’s Distance (EMD)~\citep{rubner2000earth} as the loss function to train the dynamics model:
\begin{equation}
\mathcal{L}(\mathcal{O}_t, \widehat{\mathcal{O}}_t) = \textit{w}_1 \cdot \mathcal{L}_{\text{CD}}(\mathcal{O}_t, \widehat{\mathcal{O}}_t) + \textit{w}_2 \cdot \mathcal{L}_{\text{EMD}}(\mathcal{O}_t, \widehat{\mathcal{O}}_t),
\end{equation}
where $\mathcal{O}_t$ is the real observation, and $\widehat{\mathcal{O}}_t$ is the predicted state. We select $\textit{w}_1=0.5$ and $\textit{w}_2=0.5$ based on the experiment results.
To be more specific, the CD between $\mathcal{O}_t, \widehat{\mathcal{O}}_t \subseteq \mathbb{R}^3$ is calculated by
\begin{equation}
\mathcal{L}_{\text{CD}}(\mathcal{O}_t, \widehat{\mathcal{O}}_t) =  \sum_{\mathbf{x} \in \mathcal{O}_t} \min_{\mathbf{y} \in \widehat{\mathcal{O}}_t} {\Vert \mathbf{x} - \mathbf{y} \Vert}_2^2 + \sum_{\mathbf{y} \in \widehat{\mathcal{O}}_t} \min_{\mathbf{x} \in \mathcal{O}_t} {\Vert \mathbf{x} - \mathbf{y} \Vert}_2^2.
\end{equation}
The EMD matches distributions of point clouds by finding a bijection $\mu: \mathcal{O}_t \rightarrow \widehat{\mathcal{O}}_t$ such that
\begin{equation}
\mathcal{L}_{\text{EMD}}(\mathcal{O}_t, \widehat{\mathcal{O}}_t) = \min_{\mu:\mathcal{O}_t \rightarrow \widehat{\mathcal{O}}_t} \sum_{\mathbf{x} \in \mathcal{O}_t} \|\mathbf{x}-\mu(\mathbf{x})\|_2.
\end{equation} 
We train the model to predict multiple time steps forward to regularize training and stabilize long-horizon future predictions. The training loss is calculated as a sum of distances between predictions and ground-truth states:
\begin{equation}
\mathcal{L}_{\text{train}} = \sum_{i=0}^{s} \mathcal{L}(\mathcal{O}_{t+i}, \widehat{\mathcal{O}}_{t+i}),
\end{equation} 
where the dynamics model takes $\widehat{\mathcal{O}}_{t+i-1}$ as the input to predict $\widehat{\mathcal{O}}_{t+i}$ when $i>0$. The model performs better in inference time by predicting a slightly longer horizon during training time. Empirically, we discover that $s=2$ is good enough for inference-time prediction accuracy, and a larger $s$ does not give us much gain on the accuracy. Note that although we use $s=2$ during training, we predict 15 steps during inference time, which is a long-horizon prediction. This highlights the inductive bias of our GNN-based dynamics model - we only need to train on short-horizon predictions to generalize to a much longer-horizon prediction during inference. Another motivation to keep $s$ small is to increase the training speed. More implementation details on model training can be found in Section~\ref{model_training} of supplementary materials.




\subsection{Closed-Loop Control}

\begin{figure}[t!]
\centering
\includegraphics[width=\columnwidth]{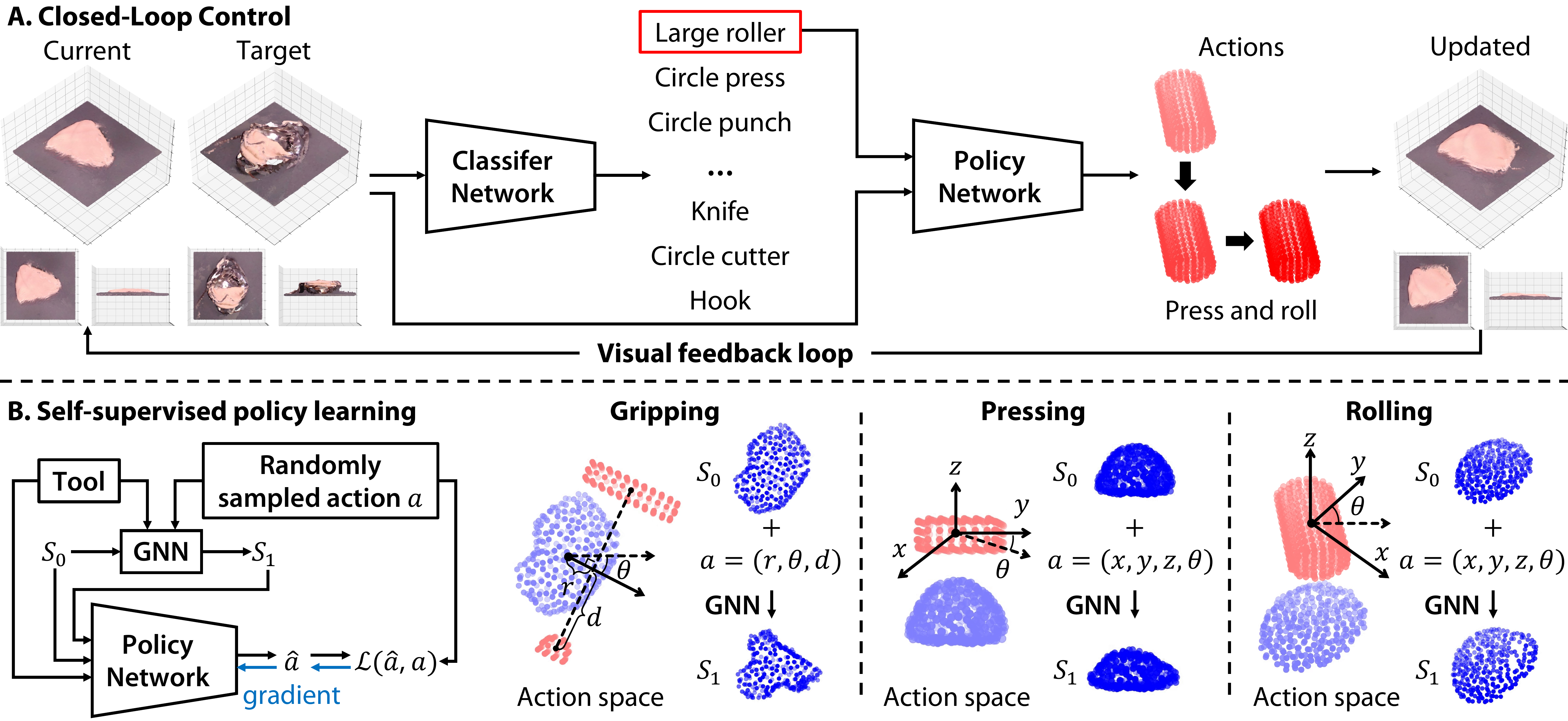}
\caption{\textbf{Planning of \model.}~\textbf{(A)} PointNet-based classifier network identifies appropriate tools based on the current observation and the target dough configuration. The self-supervised policy network takes the tool class, the current observation, and the target dough configuration as inputs and outputs the manipulation actions. The framework closes the control loop with visual feedback. \textbf{(B)} We show the policy network architecture, the parametrized action spaces of gripping, pressing, and rolling, and how we generate the synthetic datasets to train the policy network.}
\label{fig:planning}
\vspace{-2mm}
\end{figure}

To solve long-horizon manipulation tasks involving multiple tools, the robot must determine (1) the best tool to use at each stage and (2) the optimal trajectory for the selected tool.

To answer the first question, we implement a tool classification module based on PointNet++~\citep{qi2017pointnet++} that takes a concatenation of the input and output states of the dough as input and outputs probabilities of the 15 tools achieving the target state. We extract the features of the input and output point clouds separately with a PointNet++ classification layer. Then we feed the concatenated features to an MLP with a SoftMax activation layer to output probabilities.

The three tools with the highest probabilities are selected as candidates, and their optimal actions are planned. The robot selects the tool yielding the final state closest to the target and executes the planned actions. We let the module output three tool candidates instead of only one to increase our pipeline’s error tolerance to the tool classifier.

To address the second question, we specify and parameterize the action space of the 15 tools based on human priors. Then we classify the tools into a few categories based on $\left|\mathcal{A}\right|$, the dimension of their action space. Section~\ref{action_space} of supplementary materials provides more details on the specification and justification of tool action spaces.

For grippers, rollers, presses, and punches, we collect data in the real world by random exploration in the bounded action space and learn their interaction model with the dough using a GNN. Then we generate a synthetic dataset with the dynamics model, train a self-supervised policy network to obtain the optimal policy, and execute closed-loop control with visual feedback. We use straightforward planners for other tools based on human priors. More implementation details can be found in Section~\ref{action_space} of supplementary materials.

To generate the synthetic dataset, we reuse the initial dough states acquired during data collection and randomly sample actions in the parameterized action space of each tool. As shown in Figure~\ref{fig:planning}(A), the dough states before and after applying actions are the input, and the action parameters are the labels. We design a self-supervised policy network based on PointNet++. Given the continuous solution space of the action parameters, we formulate a multi-bin classification problem inspired by previous works on 3D bounding box estimation~\citep{mousavian20173d, kundu20183d}. First, we concatenate input and output point clouds, labeling them 0 or 1 for distinction. Next, we extract features using a PointNet++ layer. Finally, we feed the feature vector into separate classification and regression heads, both being MLPs with identical structures. The learned policy drastically improves the planning efficiency, and the complete pipeline only takes around 10 seconds of planning time to make a dumpling. More implementation details are described in Section~\ref{multi_bin_classification} of supplementary materials.

During closed-loop control, cameras capture the workspace point cloud, which the perception module processes to obtain a clean and sparse point cloud of the dough. Our method takes the current observation and final target as input and returns the best tool and optimal actions, as shown in Figure~\ref{fig:planning}(B). Then the robot picks up the selected tool, manipulates the dough, lifts its hand to avoid occluding the camera views, and plans the next move with visual feedback.


\section{Experiments}

\begin{figure}[t!]
\centering
\includegraphics[width=\columnwidth]{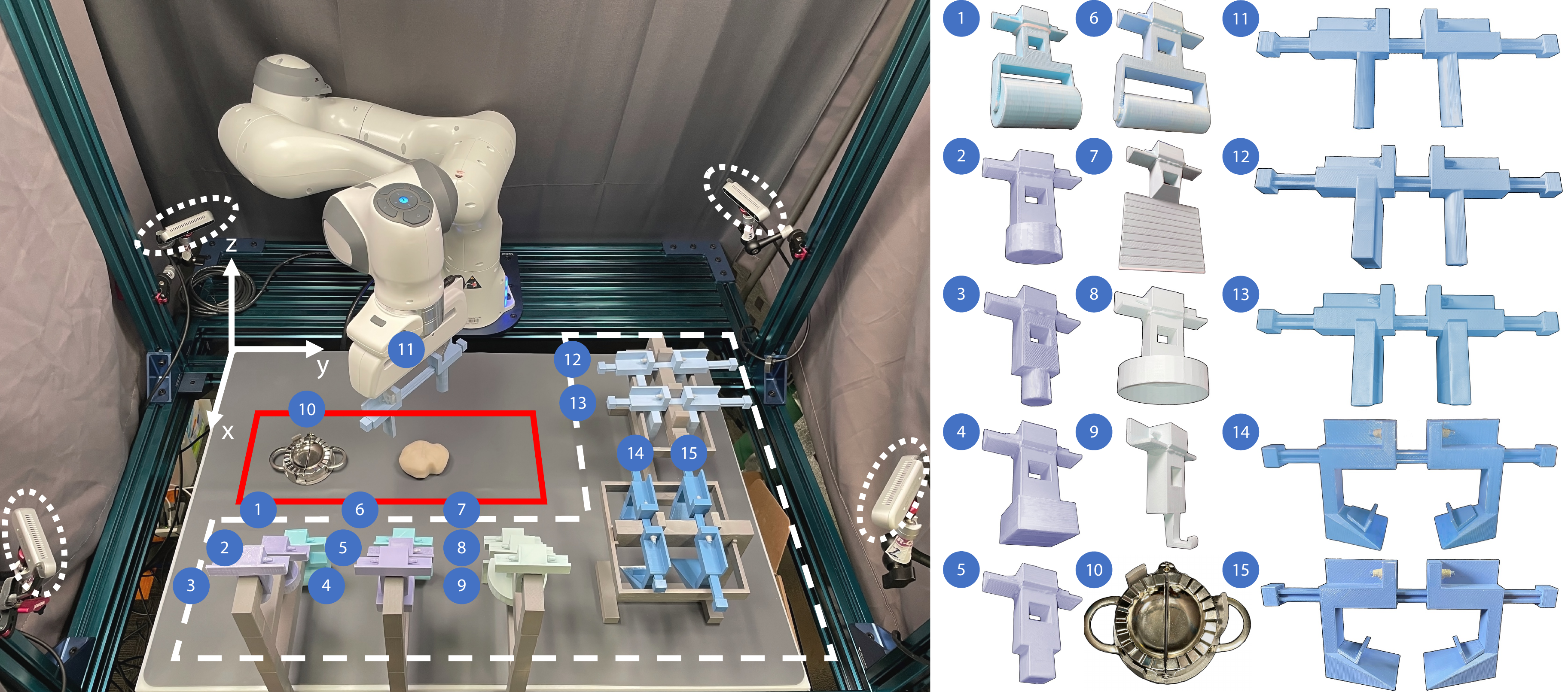}
\caption{\textbf{\model hardware and setup.}~Left: Robot's tabletop workspace with $xyz$ coordinates at top-left. Dashed white circles: four RGB-D cameras mounted at four corners of the table. Red square: dough location and manipulation area. Dashed white square: tool racks. Right: 15 tools: (1)~large roller, (2)~circle press, (3)~circle punch, (4)~square press, (5)~square punch, (6)~small roller, (7)~knife/pusher, (8)~circle cutter, (9)~hook, (10)~dumpling mold, (11)~two-rod symmetric gripper, (12)~asymmetric gripper, (13)~two-plane symmetric gripper, (14)~skin spatula, (15)~filling spatula. Tools are 3D-printed, representing common dough manipulation tools. Section~\ref{tool_design} of supplementary materials discusses the design principles of these 3D-printed tools.}
\label{fig:setup}
\vspace{-3mm}
\end{figure}

In this study, we design and implement a hardware setup for soft body manipulation tasks, allowing easy selection and interchange of 15 tools, as shown in Figure~\ref{fig:setup}. We demonstrate the \model framework's effectiveness in long-horizon soft object manipulation tasks that require the use of multiple tools, such as making a dumpling from a randomly shaped dough and shaping alphabet letter cookies to compose the word `\model.’ Additionally, \model can complete these tasks under extensive human interference, including significant changes to the shape and volume, as shown in the second supplementary video, demonstrating its robustness to external perturbations. We discuss the experiment setup in Section~\ref{setup} of supplementary materials. 

\subsection{Making Dumplings}

The dumpling-making task is to manipulate a piece of dough and the given filling into a dumpling shape. The main challenge lies in choosing appropriate tools and planning effective action trajectories. We consider the task successful if the dumpling skin is thin enough and completely covers the filling. Dumpling-making is a highly challenging task---even a single error might break the entire pipeline. Our method reliably accomplishes the task as shown in Figure~\ref{fig:dumpling}. We show a comparison with manipulation results of human subjects in Section~\ref{human_baseline} of supplementary materials.


\model also demonstrates robustness against external perturbations during real-time execution. At each stage of dumpling making, a human disturbs the robot by deforming the dough, adding additional dough, or even replacing the dough with a completely different piece to deviate from the trajectory of the robot’s original plan. For example, 
at the rolling stage, the human folds the flattened dough into a bulky shape. The robot decides to roll again to get a flatter dumpling skin. In a more challenging case, the human replaces the round dumpling skin with a completely different dough in a highly irregular shape. After this perturbation, the robot puts down the roller, picks up the knife to start again from the beginning, and successfully makes a dumpling. The complete process is shown in the second supplementary video. We also show a quantitative evaluation of the tool classification network in Section~\ref{tool_classification} of supplementary materials.


\begin{figure}[t!]
\centering
\includegraphics[width=\columnwidth]{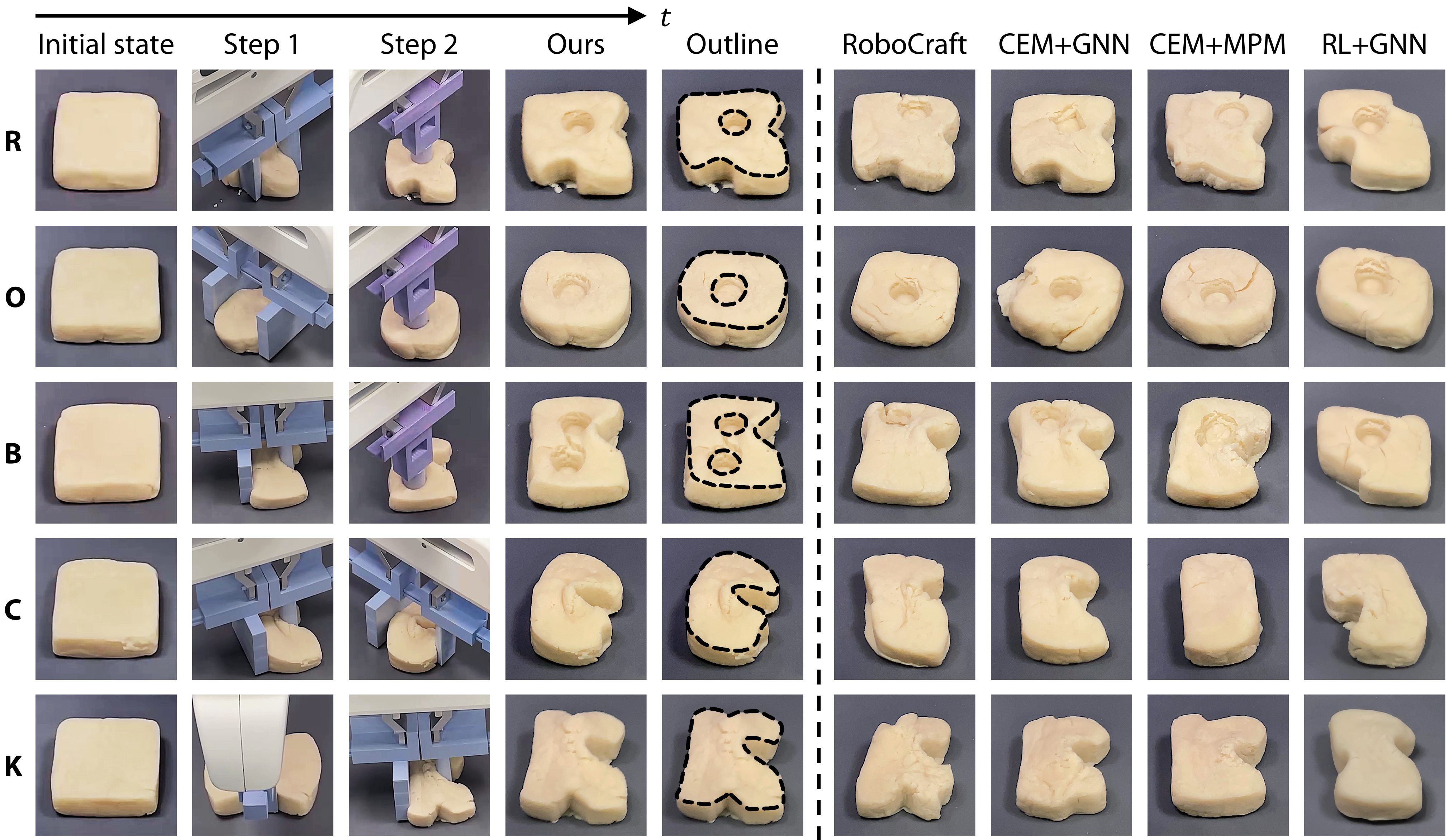}
\caption{\textbf{Making alphabetical letter cookies.}~We list R, O, B, C, and K shaping steps in Columns 1 to 4. Column 5 manually highlights the contour of the alphabetical letters. Columns 6 through 9 compare our self-supervised learned policy with three model-based planning baselines and one RL baseline. Our method can shape the dough closer to the target than all four baseline methods.}
\label{fig:letters}
\vspace*{0mm}
\end{figure}

\begin{table}[t!]
\centering
\resizebox{\textwidth}{!}{%
\begin{tabular}{lccccc}
\toprule
Methods & CD $\downarrow$ & EMD $\downarrow$ & CD of normal $\downarrow$ & Human evaluation $\uparrow$ 
& Planning time (s) $\downarrow$\\ 
\midrule
Ours & \textbf{0.0062 $\pm$ 0.0007} & \textbf{0.0042 $\pm$ 0.0006} & \textbf{0.1933 $\pm$ 0.0345} & \textbf{0.90 $\pm$ 0.11} 
& \textbf{9.3 $\pm$ 1.5} \\

RoboCraft & 0.0066 $\pm$ 0.0005 & 0.0044 $\pm$ 0.0006 & 0.2011 $\pm$ 0.0329 & 0.54 $\pm$ 0.43 
& 613.7 $\pm$ 202.7 \\

CEM+GNN & 0.0066 $\pm$ 0.0007 & 0.0045 $\pm$ 0.0008 & 0.2043 $\pm$ 0.0431 & 0.52 $\pm$ 0.41 
& 756.0 $\pm$ 234.5 \\

CEM+MPM & 0.0070 $\pm$ 0.0007 & 0.0046 $\pm$ 0.0006 & 0.1965 $\pm$ 0.0265 & 0.48 $\pm$ 0.35 & 
1486.7 $\pm$ 512.8 \\

RL+GNN & 0.0077 $\pm$ 0.0007 & 0.0064 $\pm$ 0.0009 & 0.2041 $\pm$ 0.0414 & 0.17 $\pm$ 0.09 & 
1867.9 $\pm$ 190.3 \\
\bottomrule
\end{tabular}}

\caption{\textbf{Quantitative evaluations.}~We use CD and EMD between the point clouds and the CD between the surface normals to evaluate the results.
We further profile how long these methods take to plan actions. Our method outperforms all baseline methods in these metrics by a large margin.}
\label{tab:letters}
\end{table}
\vspace*{-15mm}
\subsection{Making Alphabet Letter Cookies}

\begin{figure}[t!]
\centering
\includegraphics[width=0.9\columnwidth]{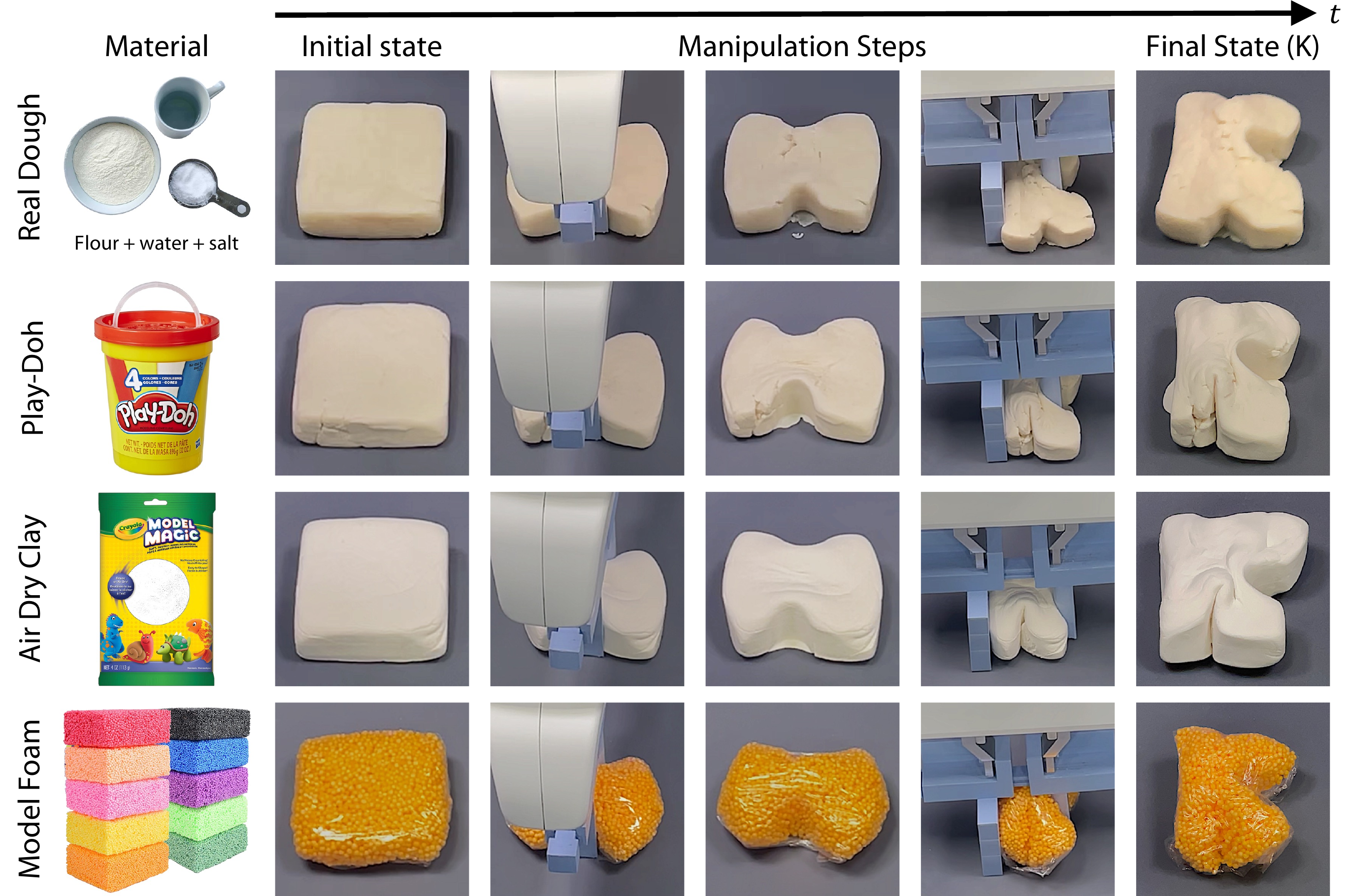}
\caption{{\textbf{Generalizing to different materials}. We showcase the dynamics model's capability to generalize to various materials by shaping a K without retraining.}}
\label{fig:materials}
\vspace{-3mm}
\end{figure}

The \model framework demonstrates effectiveness and robustness in highly complicated manipulation tasks such as dumpling-making. This section explores its generalization ability in tasks requiring precise actions such as shaping alphabet letter cookies~\citep{Shi-RSS-22} without additional training.

Figure~\ref{fig:letters} shows that the \model framework can accurately shape letters R, O, B, C, and K to compose the word ‘\model.’ For R, the robot uses the two-rod symmetric gripper for cavities and the circle punch for the hole. For O, the tool classifier selects a two-plane symmetric gripper to pinch the square into a circle and the circle punch for the hole. Our method identifies the asymmetric gripper as suitable for B and locates accurate positions for the circle punch to press twice. Shaping C is more challenging due to the large distance between the initial and target shapes, but our method successfully plans gripping positions using closed-loop visual feedback. The robot combines the two-rod symmetric and asymmetric gripper for K to create cavities.

We compare our method with four strong baselines: (1) limited-memory BFGS~\citep{fletcher2013practical} with GNN-based dynamics model (RoboCraft)~\citep{Shi-RSS-22}; (2) cross-entropy method (CEM)~\citep{rubinstein2004cross} with GNN-based dynamics model; (3) CEM with a Material Point Method (MPM) simulator~\citep{bardenhagen2004generalized}; and (4) Reinforcement Learning with GNN-based dynamics model. Qualitative results are shown in Figure~\ref{fig:letters}. We include a detailed analysis of our results in Section~\ref{analysis}.

Table~\ref{tab:letters} evaluates the results using Chamfer Distance (CD)~\citep{fan2017point}, Earth Mover’s Distance (EMD)~\citep{rubner2000earth} and CD between top surface normals. However, we recognize a discrepancy between how these metrics measure the results and how humans perceive them - these metrics are prone to local noises while humans are good at capturing the holistic features of the dough. Therefore, we show the prediction accuracy of 100 human subjects on recognizing the letters for these five methods in Table~\ref{tab:letters}. Another highlight of our method is its speed. Since the policy network only needs one forward prediction to output the actions for each tool, it is significantly faster than other methods that rely on forwarding the dynamics models many times. Section~\ref{human_evaluation} and~\ref{baseline} of supplementary materials discuss more details about the human study and the baseline implementations.

We show that the dynamics model can generalize to various materials by shaping a K without retraining in Figure~\ref{fig:materials}. We test on Play-Doh, Air Dry Clay, and Model Foam, each displaying notably different dynamics compared to our original dough made of flour, water, and salt. The dynamics model is trained solely on interaction data with the real dough.

\section{Conclusion and Limitations}
\model demonstrates its effectiveness, robustness, and generalizability in elasto-plastic object manipulation with a general-purpose robotic arm and everyday tools. The main contributions of \model include (1) tool-aware GNNs to model long-horizon soft body dynamics accurately and efficiently, (2) a tool selection module combined with dynamics models to learn tool functions through self-exploratory trials, and (3) a self-supervised policy learning framework to improve the performance and speed significantly. \model pioneers solutions for tool usage and long-horizon elasto-plastic object manipulation in building a generic cooking robot.

One limitation of \model is the occasional failure of dough sticking to the tool. A solution is to design an automatic error correction system. 
\model also relies on human priors of tool action spaces to simplify planning. But these simplifications do not constrain generalization as they can be easily specified for new tools. Section~\ref{action_space} provides more justifications for this. Another limitation is that humans define the subgoals. Higher-level temporal abstraction and task-level planning are required to get rid of them. 
Finally, \model requires additional topology estimation to apply to cables and cloths~\citep{Huang-RSS-22}, which is beyond the focus of this work. 


\clearpage
\acknowledgments{This work was in part supported by AFOSR YIP FA9550-23-1-0127, ONR MURI N00014-22-1-2740, the Toyota Research Institute (TRI), the Stanford Institute for Human-Centered AI (HAI), JPMC, and Analog Devices.}


\bibliography{references}  

\begin{thebibliography}{56}
\providecommand{\natexlab}[1]{#1}
\providecommand{\url}[1]{\texttt{#1}}
\expandafter\ifx\csname urlstyle\endcsname\relax
  \providecommand{\doi}[1]{doi: #1}\else
  \providecommand{\doi}{doi: \begingroup \urlstyle{rm}\Url}\fi

\bibitem[Matas et~al.(2018)Matas, James, and Davison]{matas2018sim}
J.~Matas, S.~James, and A.~J. Davison.
\newblock Sim-to-real reinforcement learning for deformable object
  manipulation.
\newblock In \emph{Conference on Robot Learning}, pages 734--743. PMLR, 2018.

\bibitem[Lin et~al.(2021)Lin, Wang, Olkin, and Held]{lin2021softgym}
X.~Lin, Y.~Wang, J.~Olkin, and D.~Held.
\newblock Softgym: Benchmarking deep reinforcement learning for deformable
  object manipulation.
\newblock In \emph{Conference on Robot Learning}, pages 432--448. PMLR, 2021.

\bibitem[Zhu et~al.(2022)Zhu, Cherubini, Dune, Navarro-Alarcon, Alambeigi,
  Berenson, Ficuciello, Harada, Kober, Li, et~al.]{zhu2022challenges}
J.~Zhu, A.~Cherubini, C.~Dune, D.~Navarro-Alarcon, F.~Alambeigi, D.~Berenson,
  F.~Ficuciello, K.~Harada, J.~Kober, X.~Li, et~al.
\newblock Challenges and outlook in robotic manipulation of deformable objects.
\newblock \emph{IEEE Robotics \& Automation Magazine}, 29\penalty0
  (3):\penalty0 67--77, 2022.

\bibitem[Yin et~al.(2021)Yin, Varava, and Kragic]{yin2021modeling}
H.~Yin, A.~Varava, and D.~Kragic.
\newblock Modeling, learning, perception, and control methods for deformable
  object manipulation.
\newblock \emph{Science Robotics}, 6\penalty0 (54):\penalty0 eabd8803, 2021.

\bibitem[Hartmann et~al.(2022)Hartmann, Orthey, Driess, Oguz, and
  Toussaint]{hartmann2022long}
V.~N. Hartmann, A.~Orthey, D.~Driess, O.~S. Oguz, and M.~Toussaint.
\newblock Long-horizon multi-robot rearrangement planning for construction
  assembly.
\newblock \emph{IEEE Transactions on Robotics}, 2022.

\bibitem[Nair and Finn()]{nairhierarchical}
S.~Nair and C.~Finn.
\newblock Hierarchical foresight: Self-supervised learning of long-horizon
  tasks via visual subgoal generation.
\newblock In \emph{International Conference on Learning Representations}.

\bibitem[Pirk et~al.(2021)Pirk, Hausman, Toshev, and
  Khansari]{pirk2021modeling}
S.~Pirk, K.~Hausman, A.~Toshev, and M.~Khansari.
\newblock Modeling long-horizon tasks as sequential interaction landscapes.
\newblock In \emph{Conference on Robot Learning}, pages 471--484. PMLR, 2021.

\bibitem[Simeonov et~al.(2021)Simeonov, Du, Kim, Hogan, Tenenbaum, Agrawal, and
  Rodriguez]{simeonov2021long}
A.~Simeonov, Y.~Du, B.~Kim, F.~Hogan, J.~Tenenbaum, P.~Agrawal, and
  A.~Rodriguez.
\newblock A long horizon planning framework for manipulating rigid pointcloud
  objects.
\newblock In \emph{Conference on Robot Learning}, pages 1582--1601. PMLR, 2021.

\bibitem[Billard and Kragic(2019)]{billard2019trends}
A.~Billard and D.~Kragic.
\newblock Trends and challenges in robot manipulation.
\newblock \emph{Science}, 364\penalty0 (6446):\penalty0 eaat8414, 2019.

\bibitem[Yamanobe et~al.(2017)Yamanobe, Wan, Ramirez-Alpizar, Petit, Tsuji,
  Akizuki, Hashimoto, Nagata, and Harada]{yamanobe2017brief}
N.~Yamanobe, W.~Wan, I.~G. Ramirez-Alpizar, D.~Petit, T.~Tsuji, S.~Akizuki,
  M.~Hashimoto, K.~Nagata, and K.~Harada.
\newblock A brief review of affordance in robotic manipulation research.
\newblock \emph{Advanced Robotics}, 31\penalty0 (19-20):\penalty0 1086--1101,
  2017.

\bibitem[Seita et~al.(2022)Seita, Wang, Shetty, Li, Erickson, and
  Held]{seitatoolflownet}
D.~Seita, Y.~Wang, S.~J. Shetty, E.~Y. Li, Z.~Erickson, and D.~Held.
\newblock Toolflownet: Robotic manipulation with tools via predicting tool flow
  from point clouds.
\newblock In \emph{6th Annual Conference on Robot Learning}, 2022.

\bibitem[Xie et~al.(2019)Xie, Ebert, Levine, and Finn]{Finn-RSS-19}
A.~Xie, F.~Ebert, S.~Levine, and C.~Finn.
\newblock Improvisation through physical understanding: Using novel objects as
  tools with visual foresight.
\newblock In \emph{Proceedings of Robotics: Science and Systems},
  FreiburgimBreisgau, Germany, June 2019.
\newblock \doi{10.15607/RSS.2019.XV.001}.

\bibitem[Scarselli et~al.(2008)Scarselli, Gori, Tsoi, Hagenbuchner, and
  Monfardini]{scarselli2008graph}
F.~Scarselli, M.~Gori, A.~C. Tsoi, M.~Hagenbuchner, and G.~Monfardini.
\newblock The graph neural network model.
\newblock \emph{IEEE transactions on neural networks}, 20\penalty0
  (1):\penalty0 61--80, 2008.

\bibitem[Li et~al.(2018)Li, Wu, Tedrake, Tenenbaum, and
  Torralba]{li2018learning}
Y.~Li, J.~Wu, R.~Tedrake, J.~B. Tenenbaum, and A.~Torralba.
\newblock Learning particle dynamics for manipulating rigid bodies, deformable
  objects, and fluids.
\newblock In \emph{International Conference on Learning Representations}, 2018.

\bibitem[Li et~al.(2019)Li, Wu, Zhu, Tenenbaum, Torralba, and
  Tedrake]{li2019propagation}
Y.~Li, J.~Wu, J.-Y. Zhu, J.~B. Tenenbaum, A.~Torralba, and R.~Tedrake.
\newblock Propagation networks for model-based control under partial
  observation.
\newblock In \emph{2019 International Conference on Robotics and Automation
  (ICRA)}, pages 1205--1211. IEEE, 2019.

\bibitem[Qi et~al.(2017{\natexlab{a}})Qi, Su, Mo, and Guibas]{qi2017pointnet}
C.~R. Qi, H.~Su, K.~Mo, and L.~J. Guibas.
\newblock Pointnet: Deep learning on point sets for 3d classification and
  segmentation.
\newblock In \emph{Proceedings of the IEEE conference on computer vision and
  pattern recognition}, pages 652--660, 2017{\natexlab{a}}.

\bibitem[Qi et~al.(2017{\natexlab{b}})Qi, Yi, Su, and Guibas]{qi2017pointnet++}
C.~R. Qi, L.~Yi, H.~Su, and L.~J. Guibas.
\newblock Pointnet++: Deep hierarchical feature learning on point sets in a
  metric space.
\newblock \emph{Advances in neural information processing systems}, 30,
  2017{\natexlab{b}}.

\bibitem[Shi et~al.(2022)Shi, Xu, Huang, Li, and Wu]{Shi-RSS-22}
H.~Shi, H.~Xu, Z.~Huang, Y.~Li, and J.~Wu.
\newblock {RoboCraft: Learning to See, Simulate, and Shape Elasto-Plastic
  Objects with Graph Networks}.
\newblock In \emph{Proceedings of Robotics: Science and Systems}, New York
  City, NY, USA, June 2022.
\newblock \doi{10.15607/RSS.2022.XVIII.008}.

\bibitem[Matl and Bajcsy(2021)]{matl2021deformable}
C.~Matl and R.~Bajcsy.
\newblock Deformable elasto-plastic object shaping using an elastic hand and
  model-based reinforcement learning.
\newblock In \emph{2021 IEEE/RSJ International Conference on Intelligent Robots
  and Systems (IROS)}, pages 3955--3962. IEEE, 2021.

\bibitem[Lin et~al.(2022)Lin, Qi, Zhang, Huang, Fragkiadaki, Li, Gan, and
  Held]{lin2022planning}
X.~Lin, C.~Qi, Y.~Zhang, Z.~Huang, K.~Fragkiadaki, Y.~Li, C.~Gan, and D.~Held.
\newblock Planning with spatial-temporal abstraction from point clouds for
  deformable object manipulation.
\newblock In \emph{6th Annual Conference on Robot Learning}, 2022.

\bibitem[Driess et~al.(2023)Driess, Huang, Li, Tedrake, and
  Toussaint]{driess2023learning}
D.~Driess, Z.~Huang, Y.~Li, R.~Tedrake, and M.~Toussaint.
\newblock Learning multi-object dynamics with compositional neural radiance
  fields.
\newblock In \emph{Conference on Robot Learning}, pages 1755--1768. PMLR, 2023.

\bibitem[Chi et~al.(2022)Chi, Burchfiel, Cousineau, Feng, and Song]{Chi-RSS-22}
C.~Chi, B.~Burchfiel, E.~Cousineau, S.~Feng, and S.~Song.
\newblock {Iterative Residual Policy for Goal-Conditioned Dynamic Manipulation
  of Deformable Objects}.
\newblock In \emph{Proceedings of Robotics: Science and Systems}, New York
  City, NY, USA, June 2022.
\newblock \doi{10.15607/RSS.2022.XVIII.016}.

\bibitem[Ha and Song(2022)]{ha2022flingbot}
H.~Ha and S.~Song.
\newblock Flingbot: The unreasonable effectiveness of dynamic manipulation for
  cloth unfolding.
\newblock In \emph{Conference on Robot Learning}, pages 24--33. PMLR, 2022.

\bibitem[Huang et~al.(2020)Huang, Hu, Du, Zhou, Su, Tenenbaum, and
  Gan]{huang2020plasticinelab}
Z.~Huang, Y.~Hu, T.~Du, S.~Zhou, H.~Su, J.~B. Tenenbaum, and C.~Gan.
\newblock Plasticinelab: A soft-body manipulation benchmark with differentiable
  physics.
\newblock In \emph{International Conference on Learning Representations}, 2020.

\bibitem[Lin et~al.(2021)Lin, Huang, Li, Tenenbaum, Held, and
  Gan]{lin2021diffskill}
X.~Lin, Z.~Huang, Y.~Li, J.~B. Tenenbaum, D.~Held, and C.~Gan.
\newblock Diffskill: Skill abstraction from differentiable physics for
  deformable object manipulations with tools.
\newblock In \emph{International Conference on Learning Representations}, 2021.

\bibitem[Cretu et~al.(2011)Cretu, Payeur, and Petriu]{cretu2011soft}
A.-M. Cretu, P.~Payeur, and E.~M. Petriu.
\newblock Soft object deformation monitoring and learning for model-based
  robotic hand manipulation.
\newblock \emph{IEEE Transactions on Systems, Man, and Cybernetics, Part B
  (Cybernetics)}, 42\penalty0 (3):\penalty0 740--753, 2011.

\bibitem[Li et~al.(2021)Li, Huang, Du, Su, Tenenbaum, and Gan]{li2021contact}
S.~Li, Z.~Huang, T.~Du, H.~Su, J.~B. Tenenbaum, and C.~Gan.
\newblock Contact points discovery for soft-body manipulations with
  differentiable physics.
\newblock In \emph{International Conference on Learning Representations}, 2021.

\bibitem[Navarro-Alarcon et~al.(2016)Navarro-Alarcon, Yip, Wang, Liu, Zhong,
  Zhang, and Li]{navarro2016automatic}
D.~Navarro-Alarcon, H.~M. Yip, Z.~Wang, Y.-H. Liu, F.~Zhong, T.~Zhang, and
  P.~Li.
\newblock Automatic 3-d manipulation of soft objects by robotic arms with an
  adaptive deformation model.
\newblock \emph{IEEE Transactions on Robotics}, 32\penalty0 (2):\penalty0
  429--441, 2016.

\bibitem[Cherubini et~al.(2020)Cherubini, Ortenzi, Cosgun, Lee, and
  Corke]{cherubini2020model}
A.~Cherubini, V.~Ortenzi, A.~Cosgun, R.~Lee, and P.~Corke.
\newblock Model-free vision-based shaping of deformable plastic materials.
\newblock \emph{The International Journal of Robotics Research}, 39\penalty0
  (14):\penalty0 1739--1759, 2020.

\bibitem[Yoshimoto et~al.(2011)Yoshimoto, Higashimori, Tadakuma, and
  Kaneko]{yoshimoto2011active}
K.~Yoshimoto, M.~Higashimori, K.~Tadakuma, and M.~Kaneko.
\newblock Active outline shaping of a rheological object based on plastic
  deformation distribution.
\newblock In \emph{2011 IEEE/RSJ International Conference on Intelligent Robots
  and Systems}, pages 1386--1391. IEEE, 2011.

\bibitem[Balaguer and Carpin(2011)]{balaguer2011combining}
B.~Balaguer and S.~Carpin.
\newblock Combining imitation and reinforcement learning to fold deformable
  planar objects.
\newblock In \emph{2011 IEEE/RSJ International Conference on Intelligent Robots
  and Systems}, pages 1405--1412. IEEE, 2011.

\bibitem[Nadon et~al.(2018)Nadon, Valencia, and Payeur]{nadon2018multi}
F.~Nadon, A.~J. Valencia, and P.~Payeur.
\newblock Multi-modal sensing and robotic manipulation of non-rigid objects: A
  survey.
\newblock \emph{Robotics}, 7\penalty0 (4):\penalty0 74, 2018.

\bibitem[Jia et~al.(2019)Jia, Pan, Hu, Pan, and Manocha]{jia2019cloth}
B.~Jia, Z.~Pan, Z.~Hu, J.~Pan, and D.~Manocha.
\newblock Cloth manipulation using random-forest-based imitation learning.
\newblock \emph{IEEE Robotics and Automation Letters}, 4\penalty0 (2):\penalty0
  2086--2093, 2019.

\bibitem[Figueroa et~al.(2016)Figueroa, Ureche, and
  Billard]{figueroa2016learning}
N.~Figueroa, A.~L.~P. Ureche, and A.~Billard.
\newblock Learning complex sequential tasks from demonstration: A pizza dough
  rolling case study.
\newblock In \emph{2016 11th ACM/IEEE International Conference on Human-Robot
  Interaction (HRI)}, pages 611--612. Ieee, 2016.

\bibitem[Battaglia et~al.(2016)Battaglia, Pascanu, Lai, Jimenez~Rezende,
  et~al.]{battaglia2016interaction}
P.~Battaglia, R.~Pascanu, M.~Lai, D.~Jimenez~Rezende, et~al.
\newblock Interaction networks for learning about objects, relations and
  physics.
\newblock \emph{Advances in neural information processing systems}, 29, 2016.

\bibitem[Chang et~al.(2016)Chang, Ullman, Torralba, and
  Tenenbaum]{chang2016compositional}
M.~B. Chang, T.~Ullman, A.~Torralba, and J.~B. Tenenbaum.
\newblock A compositional object-based approach to learning physical dynamics.
\newblock \emph{arXiv preprint arXiv:1612.00341}, 2016.

\bibitem[Sanchez-Gonzalez et~al.(2018)Sanchez-Gonzalez, Heess, Springenberg,
  Merel, Riedmiller, Hadsell, and Battaglia]{sanchez2018graph}
A.~Sanchez-Gonzalez, N.~Heess, J.~T. Springenberg, J.~Merel, M.~Riedmiller,
  R.~Hadsell, and P.~Battaglia.
\newblock Graph networks as learnable physics engines for inference and
  control.
\newblock In \emph{International Conference on Machine Learning}, pages
  4470--4479. PMLR, 2018.

\bibitem[Silver et~al.(2021)Silver, Chitnis, Curtis, Tenenbaum,
  Lozano-P{\'e}rez, and Kaelbling]{silver2021planning}
T.~Silver, R.~Chitnis, A.~Curtis, J.~B. Tenenbaum, T.~Lozano-P{\'e}rez, and
  L.~P. Kaelbling.
\newblock Planning with learned object importance in large problem instances
  using graph neural networks.
\newblock In \emph{Proceedings of the AAAI conference on artificial
  intelligence}, volume~35, pages 11962--11971, 2021.

\bibitem[Van~Schaik and Pradhan(2003)]{van2003model}
C.~P. Van~Schaik and G.~R. Pradhan.
\newblock A model for tool-use traditions in primates: implications for the
  coevolution of culture and cognition.
\newblock \emph{Journal of Human Evolution}, 44\penalty0 (6):\penalty0
  645--664, 2003.

\bibitem[Holladay et~al.(2019)Holladay, Lozano-P{\'e}rez, and
  Rodriguez]{holladay2019force}
R.~Holladay, T.~Lozano-P{\'e}rez, and A.~Rodriguez.
\newblock Force-and-motion constrained planning for tool use.
\newblock In \emph{2019 IEEE/RSJ International Conference on Intelligent Robots
  and Systems (IROS)}, pages 7409--7416. IEEE, 2019.

\bibitem[Parker and Gibson(1977)]{parker1977object}
S.~T. Parker and K.~R. Gibson.
\newblock Object manipulation, tool use and sensorimotor intelligence as
  feeding adaptations in cebus monkeys and great apes.
\newblock \emph{Journal of Human Evolution}, 6\penalty0 (7):\penalty0 623--641,
  1977.

\bibitem[Ingold(1993)]{ingold1993tool}
T.~Ingold.
\newblock Tool-use, sociality and intelligence.
\newblock \emph{Tools, language and cognition in human evolution}, 429\penalty0
  (45):\penalty0 449--72, 1993.

\bibitem[Matsuzawa(2008)]{matsuzawa2008primate}
T.~Matsuzawa.
\newblock Primate foundations of human intelligence: a view of tool use in
  nonhuman primates and fossil hominids.
\newblock In \emph{Primate origins of human cognition and behavior}, pages
  3--25. Springer, 2008.

\bibitem[Liang et~al.(2022)Liang, Wen, Bekris, and
  Boularias]{liang2022learning}
J.~Liang, B.~Wen, K.~Bekris, and A.~Boularias.
\newblock Learning sensorimotor primitives of sequential manipulation tasks
  from visual demonstrations.
\newblock In \emph{2022 International Conference on Robotics and Automation
  (ICRA)}, pages 8591--8597. IEEE, 2022.

\bibitem[Kazhdan et~al.(2006)Kazhdan, Bolitho, and Hoppe]{kazhdan2006poisson}
M.~Kazhdan, M.~Bolitho, and H.~Hoppe.
\newblock Poisson surface reconstruction.
\newblock In \emph{Proceedings of the fourth Eurographics symposium on Geometry
  processing}, volume~7, 2006.

\bibitem[Edelsbrunner and M{\"u}cke(1994)]{edelsbrunner1994three}
H.~Edelsbrunner and E.~P. M{\"u}cke.
\newblock Three-dimensional alpha shapes.
\newblock \emph{ACM Transactions on Graphics (TOG)}, 13\penalty0 (1):\penalty0
  43--72, 1994.

\bibitem[Attene(2010)]{attene2010lightweight}
M.~Attene.
\newblock A lightweight approach to repairing digitized polygon meshes.
\newblock \emph{The visual computer}, 26\penalty0 (11):\penalty0 1393--1406,
  2010.

\bibitem[Yuksel(2015)]{yuksel2015sample}
C.~Yuksel.
\newblock Sample elimination for generating poisson disk sample sets.
\newblock \emph{Computer Graphics Forum}, 34\penalty0 (2):\penalty0 25--32,
  2015.

\bibitem[Fan et~al.(2017)Fan, Su, and Guibas]{fan2017point}
H.~Fan, H.~Su, and L.~J. Guibas.
\newblock A point set generation network for 3d object reconstruction from a
  single image.
\newblock In \emph{Proceedings of the IEEE conference on computer vision and
  pattern recognition}, pages 605--613, 2017.

\bibitem[Rubner et~al.(2000)Rubner, Tomasi, and Guibas]{rubner2000earth}
Y.~Rubner, C.~Tomasi, and L.~J. Guibas.
\newblock The earth mover's distance as a metric for image retrieval.
\newblock \emph{International journal of computer vision}, 40\penalty0
  (2):\penalty0 99--121, 2000.

\bibitem[Mousavian et~al.(2017)Mousavian, Anguelov, Flynn, and
  Kosecka]{mousavian20173d}
A.~Mousavian, D.~Anguelov, J.~Flynn, and J.~Kosecka.
\newblock 3d bounding box estimation using deep learning and geometry.
\newblock In \emph{Proceedings of the IEEE conference on Computer Vision and
  Pattern Recognition}, pages 7074--7082, 2017.

\bibitem[Kundu et~al.(2018)Kundu, Li, and Rehg]{kundu20183d}
A.~Kundu, Y.~Li, and J.~M. Rehg.
\newblock 3d-rcnn: Instance-level 3d object reconstruction via
  render-and-compare.
\newblock In \emph{Proceedings of the IEEE conference on computer vision and
  pattern recognition}, pages 3559--3568, 2018.

\bibitem[Fletcher(2013)]{fletcher2013practical}
R.~Fletcher.
\newblock \emph{Practical methods of optimization}.
\newblock John Wiley \& Sons, 2013.

\bibitem[Rubinstein and Kroese(2004)]{rubinstein2004cross}
R.~Y. Rubinstein and D.~P. Kroese.
\newblock \emph{The cross-entropy method: a unified approach to combinatorial
  optimization, Monte-Carlo simulation, and machine learning}, volume 133.
\newblock Springer, 2004.

\bibitem[Bardenhagen and Kober(2004)]{bardenhagen2004generalized}
S.~G. Bardenhagen and E.~M. Kober.
\newblock The generalized interpolation material point method.
\newblock \emph{Computer Modeling in Engineering and Sciences}, 5\penalty0
  (6):\penalty0 477--496, 2004.

\bibitem[Huang et~al.(2022)Huang, Lin, and Held]{Huang-RSS-22}
Z.~Huang, X.~Lin, and D.~Held.
\newblock {Mesh-based Dynamics with Occlusion Reasoning for Cloth
  Manipulation}.
\newblock In \emph{Proceedings of Robotics: Science and Systems}, New York
  City, NY, USA, June 2022.
\newblock \doi{10.15607/RSS.2022.XVIII.011}.

\end{thebibliography}

\clearpage



\section{Supplementary Materials}
\etocsettocstyle{}{} 
\localtableofcontents 


\subsection{Inputs and Outputs of \model's Respective Modules}
\textbf{Perception module:} At each time step, the input is a raw observation point cloud merged from four RGBD cameras and the franka panda end-effector pose. The output is a clean and sparse surface point cloud representing the dough concatenated with a point cloud sampled on the tool mesh surface representing the action. We compute the pose of the tool point cloud from the end effector pose and the tool geometry.

\textbf{GNN dynamics:} Our GNN predicts $S_{t+1}$ from $S_t$ and $a_{t+1}$ at each prediction step. The inference-time input is the initial state $S_0$ and $\{a_t, t = 1, 2, … N\}$. The output is $S_N$. We select $N=15$ for the gripping, pressing, and rolling tasks.

\textbf{Tool predictor:} The input is a point cloud of the current dough configuration and a point cloud of the final target (e.g., the dumpling point cloud). The output is the next tool to achieve the final target configuration.

\textbf{Policy network:} The input is a point cloud of the current dough configuration and the point cloud of the subgoal associated with the selected tool. We have a one-to-one mapping between subgoals and tools. For example, a rolling pin corresponds to a dough flattened by the rolling pin. The output is the action in the defined action space of the selected tool.

\vspace{2mm}
\subsection{Perception}
\subsubsection{Data Collection} 
\label{data_collection}
To train the GNNs, we collect around 20 minutes of data for each tool using a behavior policy that randomly samples parameters within a predefined action space. In each episode, the robot applies multiple action sequences to the soft object, and after the current episode, a human manually resets the environment. Humans reshape the dough into a bulky but not necessarily cubic shape after every five grips, three rolls, and three presses.

The data collected for each tool are as follows:
(1) Asymmetric gripper~/~two-rod symmetric gripper~/~two-plane symmetric gripper: 60 episodes with five sequences per episode; (2) Circle press~/~square press~/~circle punch~/~square punch: 90 episodes with three sequences per episode; (3) Large roller~/~small roller: 80 episodes with three sequences per episode.

We collect point clouds before and after executing each sequence to train the tool classifier. However, we augment the training data by including any pair of these point clouds, not just consecutive pairs. For tools that don't require a GNN-based dynamics model, we execute a pre-coded dumpling-making pipeline ten times and add point clouds captured before and after using each tool in the pipeline to our dataset. Note that most tool selection data is a direct reuse of the dynamics data collected during the training of the dynamics model. This approach efficiently collects real-world tool selection data without needing extra exploration. We record the data collection process in the fourth supplementary video.

\subsubsection{Data Preprocessing} 
\label{data_preprocessing}
When building the dataset to train the dynamics model, aside from the sampling process of the perception module at each time frame, we also want to leverage the continuity of the video data. Therefore, we introduce simple geometric heuristics into the physical environment for better frame consistency. First, if the operating tool is not in contact with the convex hull of the object point cloud, we use the same sampled particles from the previous frame. This also applies when the tool moves away from the object. Additionally, we subsample the original videos to ensure that each video in the dataset has the same number of frames (16 frames in practice).

\subsection{Dynamics Model}
\subsubsection{Graph Building}
\label{graph_building}
When building the graph, the edges between the dough particles are constructed by finding the nearest neighbors of each particle within a fixed radius (in practice, 0.1 cm). The edges between the tool and dough particles are computed slightly differently. Instead of simply connecting to all the neighbors within the threshold, we limit the number of undirected edges between tool particles and the dough particles to at most four per tool particle to cut off the redundant edges in the graph neural network. 
Since all the node and edge features in the GNN are encoded in each particle’s local neighborhood, our GNN is naturally translation-invariant and therefore can accurately predict the movement of the dough regardless of its absolute location in the world frame.

\subsubsection{Model Training}
\label{model_training}
We train the model with temporal abstraction to enhance performance and inference speed. For example, when $t=0$, we train the model to predict the state of the dough at $t=3$ directly instead of $t=1$. This shortens the horizon, eases the task, and improves our model's inference speed by decreasing the number of forward passes needed for a full action sequence.

\subsubsection{Building Synthetic Datasets}
Since the synthetic dataset generated by the dynamics model is cheap, we can collect as much synthetic data as desired. Empirically, for example, we sample 128 random actions in the action space of gripping for each of the 300 different initial states, so there are $128*300=38,400$ pairs of input and output point clouds to train the policy network. The random walk to generate the synthetic dataset starts from initial dough states acquired during dynamics data collection, which are not all block-shaped states and cover various shapes.

\subsection{Closed-loop Control}
\subsubsection{Action Space}
\label{action_space}
We classify the tools into a few categories based on $\left|\mathcal{A}\right|$, the dimension of their corresponding action space. We visualize action spaces for gripping, pressing, and rolling in Figure~\ref{fig:planning} (B).
\begin{enumerate}[label=\Alph*)]
    \item Nine tools that have an action space with $\left|\mathcal{A}\right| \geq 3$: 
    \begin{enumerate}[label=\arabic*)]
        \item Asymmetric gripper~/~two-rod symmetric gripper~/~two-plane symmetric gripper: ${\left\{r,\theta,d\right\}}$, where $r$ is the distance between the midpoint of the line segment connecting the centers of mass of the gripper’s two fingers and the center of the target object, $\theta$ is the robot gripper’s rotation about the (vertical) axis, and $d$ is the minimal distance between the gripper’s two fingers during this pinch. 
        \item Large roller~/~small roller~/~square press~/~square punch: ${\left\{x, y, z, \theta\right\}}$, where ${\left\{x, y, z\right\}}$ is the bounded location indicating the center of the action, and $\theta$ is the robot gripper’s rotation about the vertical axis. In the case of rollers, the rolling distance is fixed and therefore not included in the action space.
        \item Circle press~/~circle punch: ${\left\{x, y, z\right\}}$, where ${\left\{x, y, z\right\}}$ is the bounded location indicating the center of the action. The robot gripper’s rotation is unnecessary because the tool's bottom surface is a circle.
    \end{enumerate} 
    
    \item Five tools that have an action space with $\left|\mathcal{A}\right| = 2$:
    \begin{enumerate}[label=\arabic*)]
        \item Knife~/~circle cutter~/~pusher~/~skin spatula~/~filling spatula: ${\left\{x, y\right\}}$, where ${\left\{x, y\right\}}$ is the bounded location indicating the center of the action on the plane. $\theta$ and $z$ are fixed for these tools to simplify the action space. 
    \end{enumerate}
    
    \item The action of the hook is precoded.

\end{enumerate}

In category B, for all tools except the knife, we leverage the prior that the center of the dough is always the optimal solution in the action space and directly compute the center from the processed point cloud. In the case of the knife, we use the $y$ coordinate of the center of the dough as the solution for $y$ (the $xyz$ coordinate system is illustrated in Figure~\ref{fig:setup}). For $x$, we first compute the volume of the target dough and then perform a binary search with the center of the dough as the starting point to find the cut position that results in the closest volume to the target volume.

In category C, the hook is precoded first to hook the handle of the dumpling mold, then close the mold, press the mold handle to turn the dough into a dumpling shape, and finally open the mold by hooking and lifting the handle.

The guiding principle in designing action spaces involves starting with the end-effector's 6-DoF action space and eliminating redundant DoFs. For instance, rotations along the $x$ and $y$ axes are typically not required to generate a meaningful action. Hence, we opt to exclude them from the action space of the 14 tools. For grippers, we transform the Cartesian coordinate system into a polar coordinate system to simplify the search process for action parameters since corner cases in the bounded Cartesian space are usually suboptimal. Following this, we introduce tool-specific DoFs, which are determined by the tool's geometric properties. For example, in the case of grippers, we incorporate an additional parameter, $d$, to represent the width between the gripper's two fingers.

Our method can potentially generalize to various challenging dough manipulation tasks besides dumpling-making, such as making alphabet letter cookies (as shown in the paper), pizza, and noodles. A successful transfer requires the ground truth meshes of new tools and data from interacting with them. We only need 20 minutes of real-world interaction data per tool, demonstrating the ease of retraining for new tasks and tools. Although we incorporate human prior knowledge to simplify the action space for tools, it does not constrain the generalization capability since we can easily specify the action space for new tools. One limitation is that hand-designed action spaces may not be desirable in general manipulation settings. Our insight is that for most tasks, especially tasks where the robot uses a tool, there is much redundant and undesired space in the full 6-DoF action space. In most cases, humans follow certain rules and a constrained action space when using a specific tool. A future direction is to design a planner to automatically prune the action space conditioned on the tool and the task.

\subsubsection{Multi-bin Classification}
\label{multi_bin_classification}
We formulate the self-supervised policy training as a multi-bin classification problem inspired by previous works on 3D bounding box estimation~\citep{mousavian20173d, kundu20183d}. The total loss for the multi-bin classification is
\begin{align}
\mathcal{L} = \sum_{i=1}^{\left|\mathcal{A}\right|} \left(\mathcal{L}_{\text{conf}}^{\mathcal{A}_i} + \textit{w} \cdot \mathcal{L}_{\text{loc}}^{\mathcal{A}_i}\right),
\end{align} 
where the confidence loss $\mathcal{L}_{\text{conf}}^{\mathcal{A}_i}$ is the softmax loss of the confidences of each bin for each action parameter $\mathcal{A}_i$, and the localization loss $\mathcal{L}_{\text{loc}}^{\mathcal{A}_i}$ is the loss that tries to minimize the difference between the estimated parameter and the ground truth parameter. For orientation estimation, we use negative cosine loss as the localization loss and force it to minimize the difference between the ground truth and all the bins that cover that value. We use the smooth L1 loss as the localization loss for action parameters not representing an orientation. During inference time, for each parameter, the bin with maximum confidence is selected, and the final output is computed by adding the estimated delta of that bin to the center of the same bin.

To establish the bins, we first bound our action space into a reasonable range $(A_{min}, A_{max})$. Second, we define the number of bins $N$ as a hyperparameter and select $N=8$ for translations and $N=32$ for rotations. Third, we divide the action space into $N+1$ bins with size $2 * (A_{max} - A_{min}) / N$. The center of the first bin is at $A_{min}$, the center of the last bin is at $A_{max}$, and each bin overlaps with neighboring bins. This approach is similar to~\citep{mousavian20173d}.
\subsubsection{Subgoal Definitions}
As mentioned in \ref{data_collection}, during data collection, we execute a hand-coded dumpling-making pipeline ten times and add point clouds captured before and after using each tool in the pipeline to our dataset as expert demonstrations. The point clouds recorded after using each tool in one of these trajectories are selected as subgoals.

\subsection{Experiment Setup}\label{setup}
\subsubsection{Robot and Sensors}\label{robot}
We use the 7-DoF Franka Emika Panda robot arm and its parallel jaw gripper as the base robot. Four calibrated Intel RealSense D415 RGB-D cameras are fixed on vertical metal bars around the robot tabletop, as shown in Figure~\ref{fig:setup}. The cameras capture 1280$\times$720 RGB-D images at 30 Hz. We also design a set of 3D-printed tools based on real-world dough manipulation tools.

\subsubsection{Tool Design}
\label{tool_design}
\begin{figure}[t!]
\centering
\includegraphics[width=\columnwidth]{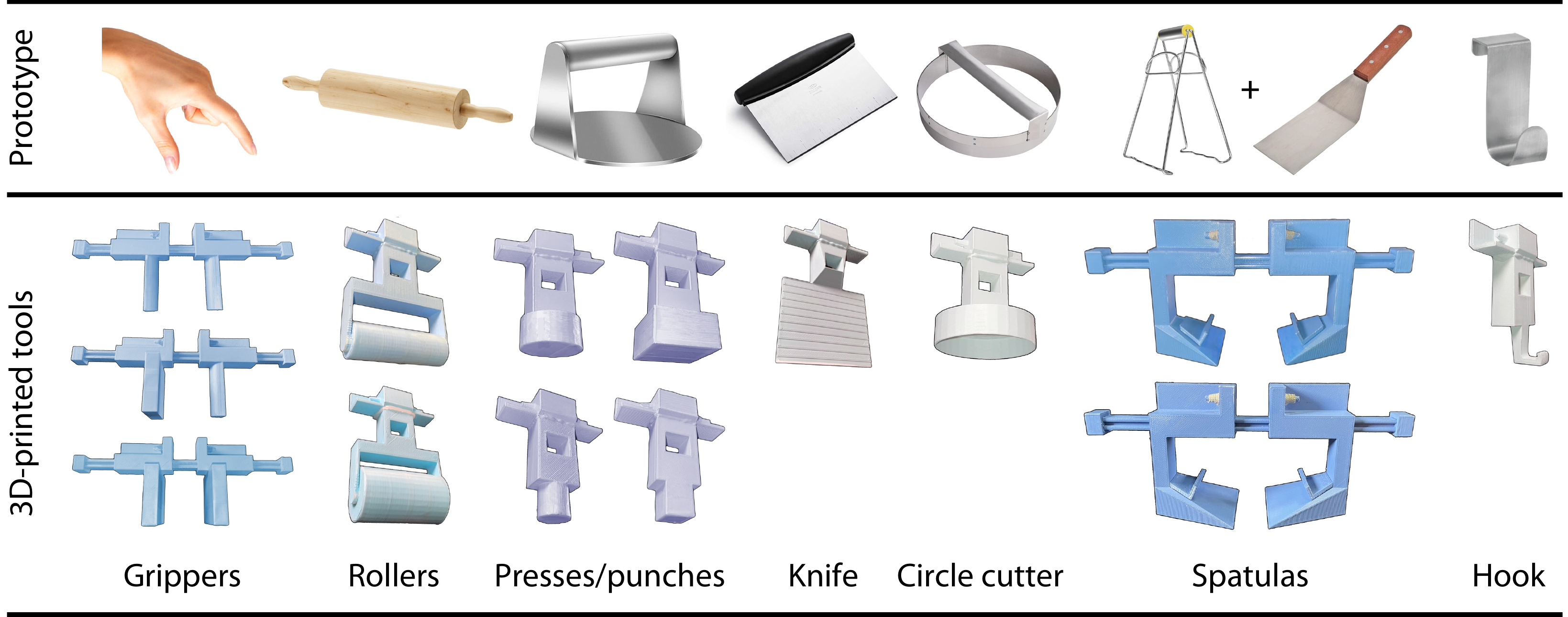}
\caption{\textbf{Prototypes of 3D-printed tools}. We show a comparison between our 3D-printed tools and their real-world prototypes which are common kitchen tools for dough manipulation. The design principle of these 3D-printed tools is to mimic real-world ones as closely as possible. We use 3D-printed tools instead of real-world ones to allow the robot arm to acquire and manipulate the tools more easily.}
\label{fig:prototype}
\vspace{-0.1em}
\end{figure}

We design and 3D-print 14 tools: large roller, small roller, circle press, circle punch, square press, square punch, knife~/~pusher, circle cutter, two-rod symmetric gripper, asymmetric gripper, two-plane symmetric gripper, skin spatula, filling spatula, and hook. The dumpling mold is the same as real-world ones. In Figure~\ref{fig:prototype}, we compare our 3D-printed tools and their real-world prototypes, which are common kitchen tools for dough manipulation. The design principle of these 3D-printed tools is to mimic real-world ones as closely as possible.

The roller is composed of a holder and a rolling pin so that the rolling pin can rotate freely while the holder remains static. We designed both large and small rollers to accommodate different needs. We also have a set of punches and presses with square and circle shapes. The knife is a thin planar tool that can cut through objects. Similarly, the circle cutter can cut an object into a circular shape. Among the grippers, the two-rod symmetric gripper consists of two cylindrical extrusions, the asymmetric gripper consists of a cylindrical and planar part, and the two-plane symmetric gripper consists of two planar parts. The two extruding rods on each gripper insert into the corresponding holes of the two fingers of Franka’s gripper, allowing them to adhere to and move along with the fingers. A linear shaft connects the two parts of each gripper, constraining their movement to a single axis. The skin and filling spatulas have a similar design, except that their two extrusions are each spatula, so they can pick up and put down the soft object without deforming it. The hook and the dumpling mold are tools used together to mold the dough into a dumpling shape.

\begin{figure}[t!]
\centering
\includegraphics[width=\columnwidth]{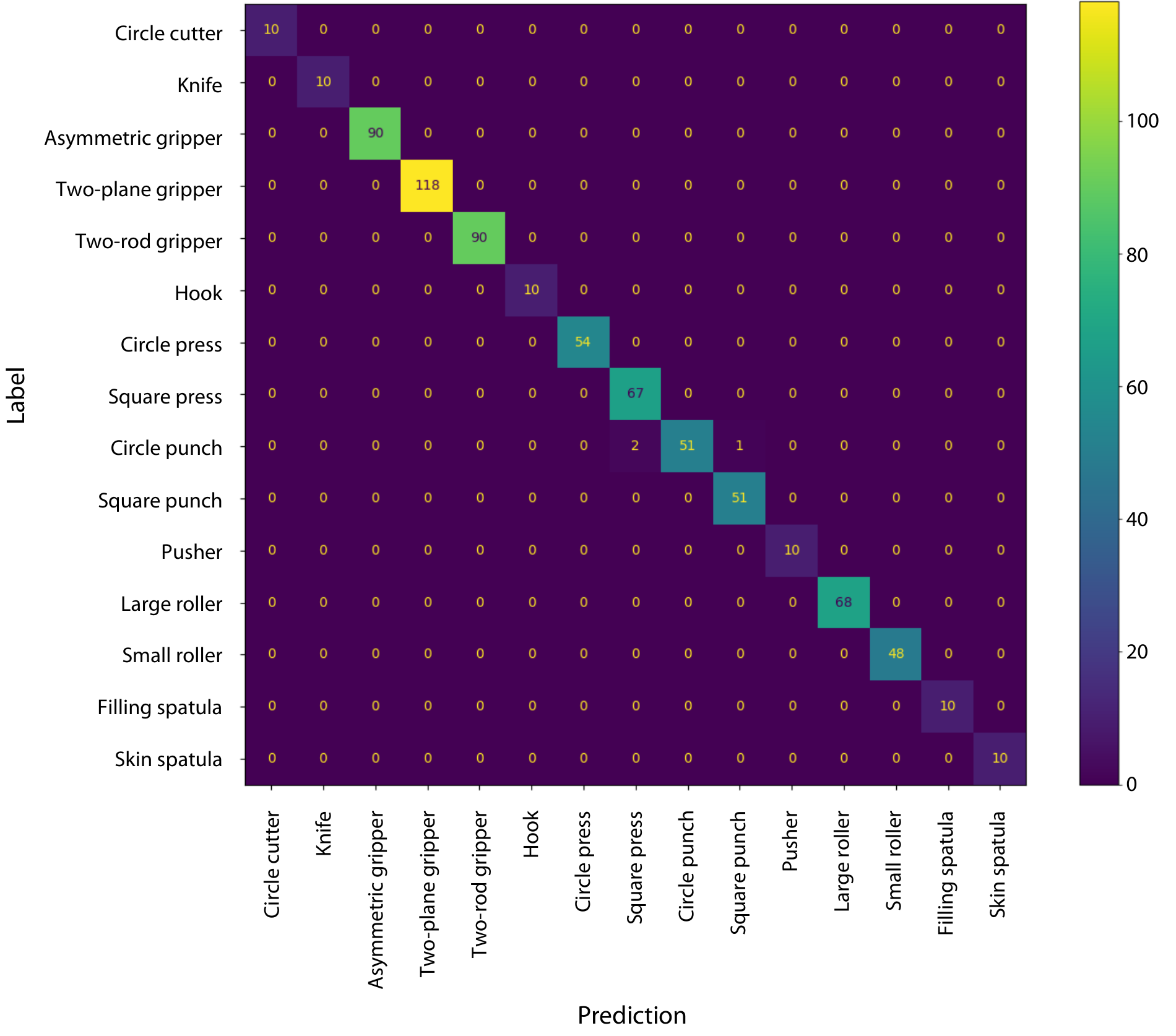}
\caption{\textbf{Confusion matrix of the tool classifier predictions}. We show the confusion matrix of the tool classifier predictions on the test set, which is split from the training data. The tool classifier achieves an accuracy very close to 1.}
\label{fig:cm}
\vspace{-0.1em}
\end{figure}

\subsubsection{Tool-Switching Setup}
\label{switching}
The tool-switching setup is an engineering innovation we implement in this project. We adopt two designs so that the robot can pick up, use, and put down the tools without any help from humans: (1) The connector on the top of each tool attaches firmly to the Franka’s gripper when it closes its fingers and also unlocks easily when the gripper reopens. (2) The tool racks on the bottom and right side of the robot table hold all the 3D-printed tools in their upright poses so that the robot can easily pick up and put down the tools. Additionally, we calibrate the tools’ world coordinates so that the robot knows where to find each tool. The supplementary videos of making dumplings show examples of how the robot switches tools.

\begin{figure}[t!]
\centering
\includegraphics[width=\columnwidth]{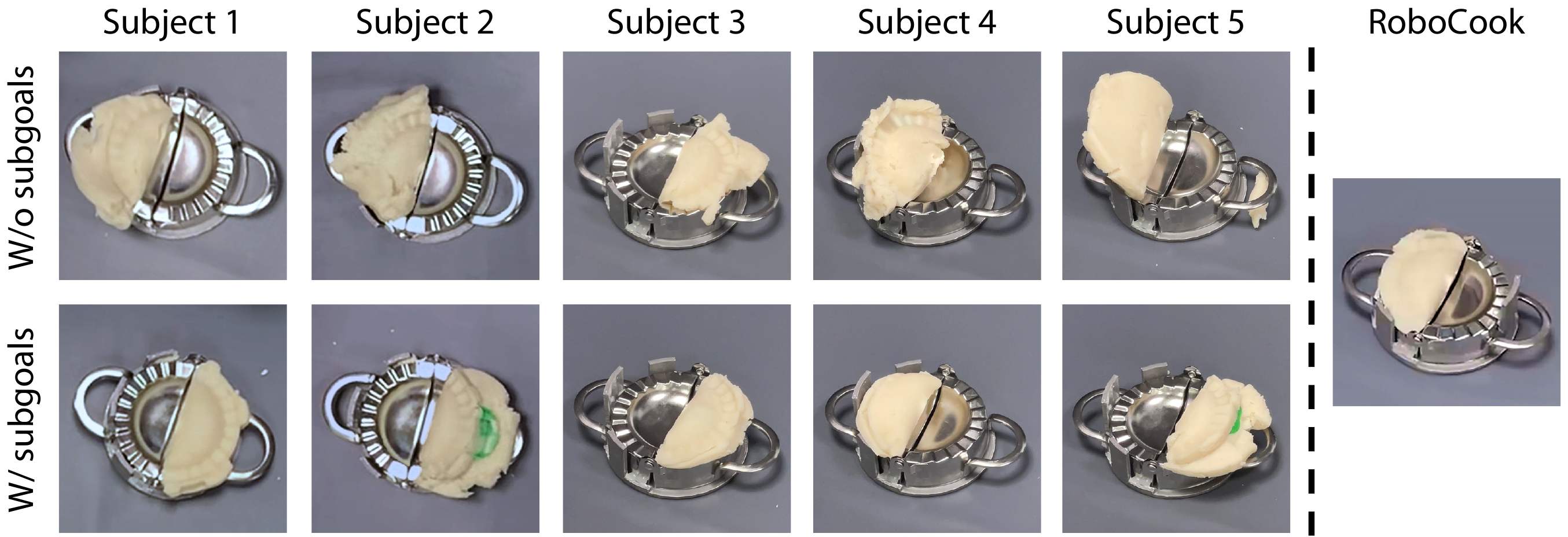}
\caption{\textbf{Comparison with human subjects}. We show a comparison with the manipulation results of human subjects. In the first row, Human subjects devise their manipulation plan and choose tools independently. In the second row, human subjects follow a given tool sequence and subgoals.}
\label{fig:human}
\vspace{-0.1em}
\end{figure}

\subsection{Comparison with Human Subjects on Dumpling-making}
\label{human_baseline}
We invited five human subjects to make dumplings with the same tools to highlight the complexity of dumpling-making. Each subject participated in two experiments: choosing tools independently and following a given tool sequence and subgoals. For a fair comparison, human subjects were not allowed to directly touch the dough with their hands or apply each tool more than five times. Before the experiments, we introduced each tool and gave them sufficient time to get familiar with the dough's dynamics and devise their plan. We compare their best attempt among three trials to our method for each experiment. Figure~\ref{fig:human} shows that human performance is notably worse than our method without subgoals. Performance improves with the tool sequence and subgoals but remains comparable to or worse than our method. The fifth supplementary video records the entire process.

\subsection{Tool Classification}
\label{tool_classification}
The training of our tool classifier is supervised learning with a cross-entropy loss as in standard classification architectures. We split a test set from the training data of the tool classifier and show the confusion matrix of the tool classifier predictions in Figure ~\ref{fig:cm}. The instance accuracy is 0.996. We compared PointNet-based and ResNet-based architectures for the tool classification network. PointNet-base architecture generalizes better due to its ability to encode depth information. Empirically, it demonstrates greater robustness to changes in lighting, dough color, and dough transformations.

\subsection{Human Evaluation of Alphabet Letters}
\label{human_evaluation}
We recognize a discrepancy between how metrics such as Chamfer Distance measure the results and how humans perceive them - these metrics are prone to local noises while humans are good at capturing the holistic features of the dough. Therefore, we invite 100 human subjects to evaluate the results. The human survey asks the question: “What alphabet letter is the robot trying to shape in the given image?” If we put all 20 images (four methods $\times$ five letters) in Question 1, there could be a predictive bias from seeing more than one image of the same letter. Therefore, we shuffle the order of 20 images and split them into four groups. Each group contains one image for each letter but from different methods. After averaging over five letters, we show the human perceived accuracy and human ranking of the performance of these four methods in Table~\ref{tab:letters}.

\subsection{Baseline Implementation Details}\label{baseline}
Baselines RoboCraft, CEM+GNN, and RL+GNN use the same framework as \model but replace the policy network with gradient descent (RoboCraft), CEM, and RL. In other words, GNNs in these baselines are the same as GNNs we trained in \model. They also use the same tool predictor as in \model to handle this multi-tool task. They are meant to compare the policy network in \model with alternatives.

The RL+GNN baseline utilizes a model-based Soft Actor-Critic (SAC) with a learned GNN-based dynamics model as the world model. The action space aligns with other planning methods, and the state space comprises the point cloud position. The reward function is derived from the change in Chamfer Distance after each grip. Training involves a discount factor of 0.99, a learning rate of 0.0003 with the Adam optimizer, 2-layer MLPs with 256 hidden units, and ReLU activation for both policy and critic models. We initially collect 250 steps of warm-up data. The replay buffer size is 1e6, and the target smooth coefficient is 0.005. We trained the RL baseline for 2500 steps online for each action prediction to fit into our closed-loop control module. The inputs and outputs of the RL baseline are the same as our policy network. The input is a point cloud of the current dough configuration and the point cloud of the subgoal. The output is the action in the defined action space of the selected tool.
The results shown in Figure~\ref{fig:letters} and Table~\ref{tab:letters} indicate that the RL baseline is noticeably worse than our method.

\subsection{Analysis of Comparison with Baselines}\label{analysis}

Our methods outperform baselines RoboCraft, CEM+GNN, and RL+GNN by a large margin, using the same GNNs and tool predictors but different planning methods. One reason is that our policy network sees a complete coverage of the entire action space from the large synthetic datasets generated by our dynamics model offline. Other planning methods explore the action space online and therefore have a smaller coverage than ours. By training a goal-conditioned policy network offline, we also make the online action synthesis much more efficient than baseline methods.

Another insight is that our PointNet-based policy network is better at capturing higher-level shape changes of the point cloud, such as concavities. For example, by comparing the current and target dough configuration, the policy network knows that the gripper should probably pinch the concave locations in the target dough configuration to deform the current dough into the target shape.

Our method also outperforms the baseline CEM+MPM since an MPM simulator suffers from a sim2real gap and relies on careful system identification to bridge the gap. Thus, the MPM simulator underperforms the GNN, which is directly trained on real-world data.


\end{document}